\def\year{2021}\relax
\documentclass[letterpaper]{article} 
\usepackage{aaai21}  
\usepackage{times}  
\usepackage{helvet} 
\usepackage{courier}  
\usepackage[hyphens]{url}  
\usepackage{graphicx} 
\usepackage{natbib}  
\usepackage{caption} 
\usepackage{latexsym}
\usepackage{makecell}
\usepackage{amsmath,amssymb,mathtools,bm,etoolbox}
\usepackage{algorithm}
\usepackage[noend]{algorithmic}
\usepackage{enumitem}
\usepackage{xcolor}
\usepackage{pifont}
\usepackage{multirow}
\usepackage{diagbox}
\usepackage[switch]{lineno}
\usepackage[autostyle=false, style=english]{csquotes}
\MakeOuterQuote{"}
\usepackage[draft]{hyperref} 

\DeclarePairedDelimiter\floor{\lfloor}{\rfloor}

\newcommand{\eg}{\emph{e.g.}}

\newcommand{\ie}{\emph{i.e.}}

\usepackage{microtype}

\title{Learning by Fixing: Solving Math Word Problems with Weak Supervision}
\author {
    Yining Hong,
    Qing Li,
    Daniel Ciao,
    Siyuan Huang,
    Song-Chun Zhu\\
}
\affiliations {
     University of California, Los Angeles, USA. \\
    yininghong@cs.ucla.edu,
    \{liqing, danielciao, huangsiyuan\}@ucla.edu,
    sczhu@stat.ucla.edu
}
\begin{document}
\maketitle

\begin{abstract}
Previous neural solvers of math word problems (MWPs) are learned with full supervision and fail to generate diverse solutions. In this paper, we address this issue by introducing a \textit{weakly-supervised} paradigm for learning MWPs. Our method only requires the annotations of the final answers and can generate various solutions for a single problem. To boost weakly-supervised learning, we propose a novel \textit{learning-by-fixing} (LBF) framework, which corrects the misperceptions of the neural network via symbolic reasoning. Specifically, for an incorrect solution tree generated by the neural network, the \textit{fixing} mechanism propagates the error from the root node to the leaf nodes and infers the most probable fix that can be executed to get the desired answer. To generate more diverse solutions, \textit{tree regularization} is applied to guide the efficient shrinkage and exploration of the solution space, and a \textit{memory buffer} is designed to track and save the discovered various fixes for each problem. Experimental results on the Math23K dataset show the proposed LBF framework significantly outperforms reinforcement learning baselines in weakly-supervised learning. Furthermore, it achieves comparable top-1 and much better top-3/5 answer accuracies than fully-supervised methods, demonstrating its strength in producing diverse solutions.
\end{abstract}

\section{Introduction}

\begin{figure}[t]
    \vspace{-0.5cm}
    \centering
    \includegraphics[width=\linewidth]{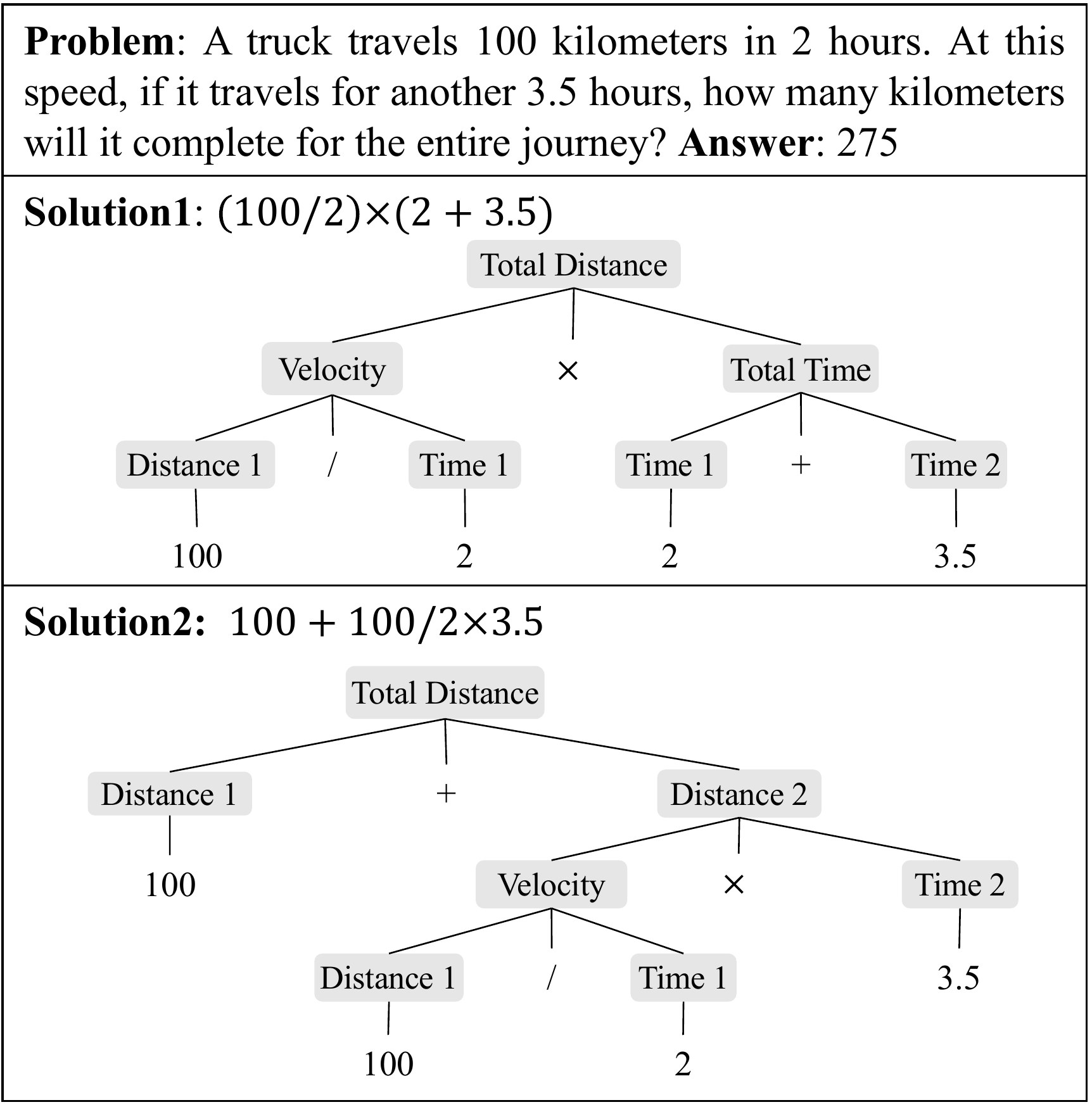}
    \caption{Exemplar MWP with multiple solutions.}
    \label{fig:example}
    \vskip -0.2in
\end{figure}

Solving math word problems (MWPs) poses unique challenges for understanding natural-language problems and performing arithmetic reasoning over quantities with commonsense knowledge. As shown in \autoref{fig:example}, a typical MWP consists of a short narrative describing a situation in the world and asking a question about an unknown quantity. To solve the MWP in \autoref{fig:example}, a machine needs to extract key quantities from the text, such as "100 kilometers" and "2 hours", and understand the relationships between them. General mathematical knowledge like "distance = velocity $\times$ time" is then used to calculate the solution. 


Researchers have recently focused on solving MWPs using neural-symbolic models~\cite{Ling2017ProgramIB, wang-etal-2017-deep, Huang2018NeuralMW, Wang2018TranslatingMW, Xie2019AGT}. These models usually consist of a neural perception module (\ie, Seq2Seq or Seq2Tree) that maps the problem text into a solution expression or tree, and a symbolic module which executes the expression and generates the final answer. Training these models requires the full supervision of the solution expressions.

However, these fully-supervised approaches have three drawbacks. First, current MWP datasets only provide one solution for each problem, while there naturally exist multiple solutions that give different paths of solving the same problem. For instance, the problem in \autoref{fig:example} can be solved by ``$(100/2) \times (2+3.5)$'' if we first calculate the speed and then multiply it by the total time; alternatively, we can solve it using ``$100 + 100/2 \times 3.5$'' by summing the distances of the first and second parts of the journey. The models trained with full supervision on current datasets are forced to fit the given solution and cannot generate diverse solutions. Second, annotating the expressions for MWPs is time-consuming. However, a large amount of MWPs with their final answers can be mined effortlessly from the internet (\eg, online forums). How to efficiently utilize these partially-labeled data without the supervision of expressions remains an open problem. Third, current supervised learning approaches suffer from the train-test discrepancy. The fully-supervised learning methods optimize expression accuracy rather than answer accuracy. However, the model is evaluated by the answer accuracy on the test set, causing a natural performance gap.

To address these issues, we propose to solve the MWPs with \textit{weak supervision}, where only the problem texts and the final answers are required. By directly optimizing the answer accuracy rather than the expression accuracy, learning with weak supervision naturally addresses the train-test discrepancy. Our model consists of a tree-structured neural model similar to \citet{Xie2019AGT} to generate the solution tree and a symbolic execution module to calculate the answer. However, the symbolic execution module for arithmetic expressions is non-differentiable with respect to the answer accuracy, making it infeasible to use back-propagation to compute gradients. A straightforward approach is to employ policy gradient methods like REINFORCE~\cite{williams1992simple} to train the neural model. The policy gradient methods explore the solution space and update the policy based on generated solutions that happen to hit the correct answer. Since the solution space is large and incorrect solutions are abandoned with zero reward, these methods usually converge slowly or fail to converge.

To improve the efficiency of weakly-supervised learning, we propose a novel \textit{fixing} mechanism to learn from incorrect predictions, which is inspired by the human ability to learn from failures via abductive reasoning~\cite{magnani2009abductive,Zhou2018AbductiveLT}. The fixing mechanism propagates the error from the root node to the leaf nodes in the solution tree and finds the most probable \textit{fix} that can generate the desired answer. The fixed solution tree is further used as a pseudo label to train the neural model. \autoref{fig:framework} shows how the fixing mechanism corrects the wrong solution tree by tracing the error in a top-down manner.

Furthermore, we design two practical techniques to traverse the solution space and discover possible solutions efficiently. First, we observe a positive correlation between the number of quantities in the text and the size of the solution tree (the number of leaf nodes in the tree), and propose a \textit{tree regularization} technique based on this observation to limit the range of possible tree sizes and shrink the solution space. Second, we adopt a \textit{memory buffer} to track and save the discovered fixes for each problem with the fixing mechanism. All memory buffer solutions are used as pseudo labels to train the model, encouraging the model to generate more diverse solutions for a single problem. 

In summary, by combining the fixing mechanism and the above two techniques, the proposed \textbf{learning-by-fixing} (LBF) method contains an exploring stage and a learning stage in each iteration, as shown in \autoref{fig:framework}. We utilize the fixing mechanism and tree regularization to correct wrong answers in the exploring stage and generate fixed expressions as pseudo labels. In the learning stage, we train the neural model using these pseudo labels.

We conduct comprehensive experiments on the Math23K dataset~\cite{wang-etal-2017-deep}. The proposed LBF method significantly outperforms the reinforcement learning baselines in weakly-supervised learning and achieves comparable performance with several fully-supervised methods. Furthermore, our proposed method achieves significantly better answer accuracies of all the top-3/5 answers than fully-supervised methods, illustrating its advantage in generating diverse solutions. The ablative experiments also demonstrate the efficacy of the designed algorithms, including the fixing mechanism, tree regularization, and memory buffer.

\section{Related Work}
\subsection{Math Word Problems}

 Recently, there emerges various question-answering tasks that require human-like reasoning abilities \cite{Qi2015ARV, Tu2014JointVA, Zhang2019RAVENAD, Dua2019DROPAR, hong2019academic, zhu2020dark, Zhang2020MachineNS,li2020competence, Yu2020ReClorAR}. Among them, solving mathematical word problems (MWPs) is a fundamental and challenging task.
 
 Previous studies of MWPs range from traditional rule-based methods~\cite{Fletcher1985UnderstandingAS, Bakman2007RobustUO, Yuhui2010FrameBasedCO}, statistical learning methods~\cite{Kushman2014LearningTA, Zhou2015LearnTS, Mitra2016LearningTU,  Roy2017UnitDG, Huang2016HowWD}, semantic-parsing methods~\cite{Shi2015AutomaticallySN, KoncelKedziorski2015ParsingAW, Huang2017LearningFE} to recent deep learning methods~\cite{Ling2017ProgramIB, wang-etal-2017-deep, Huang2018NeuralMW, Robaidek2018DataDrivenMF, Wang2018TranslatingMW, Wang_Zhang_Zhang_Xu_Gao_Dai_Shen_2019, Chiang2019SemanticallyAlignedEG, Xie2019AGT, zhang2020graph2tree}. 
 
 In particular, Deep Neural Solver (DNS)~\cite{wang-etal-2017-deep} is a pioneering work that designs a Seq2seq model to solve MWPs and achieves promising results. \citet{Xie2019AGT} propose a tree-structured neural solver to generate the solution tree in a goal-driven manner. All these neural solvers learn the model with full supervision, where the ground-truth intermediate representations (e.g., expressions, programs) are given during training. To learn the solver with less supervision, \citet{KoncelKedziorski2015ParsingAW} use a discriminative model to solve MWPs in a weakly-supervised way. They utilize separate modules to extract features, construct expression trees, and score the likelihood, which is different from the current end-to-end neural solvers. \citet{Upadhyay2016LearningFE}, \citet{Zhou2015LearnTS}, and \citet{Kushman2014LearningTA} use mixed supervision, where one dataset has only annotated equations,  and the other has only final answers. However, for the set with final answers, they also depend on pre-defined equation templates. \citet{Chen2020NeuralSR} apply a neural-symbolic reader on MathQA\cite{Amini2019MathQATI}, which is a large-scale dataset  with fully-specified  operational  programs. They have access to the ground truth programs for a small fraction of training samples at the first iterations of training.
Unlike these methods, the proposed LBF method requires only the supervision of the final answer and generates diverse solutions by keeping a memory buffer. Notably, it addresses the sparse reward problem in policy gradient methods using a fixing mechanism that propagates error down a solution tree and finds the most probable fix.


\subsection{Neural-Symbolic Learning for NLP}
Neural-symbolic learning has been applied to solve NLP tasks with weak supervision, such as semantic parsing and program synthesis~\cite{liang2016neural, Guu2017FromLT, Liang2018MemoryAP, Agarwal2019LearningTG,li2020competence}. Similar to MWP, they generate intermediate symbolic representations with a neural network and execute the intermediate representation with a symbolic reasoning module to get the final result. Typical approaches for such neural-symbolic models use policy gradient methods like REINFORCE since the symbolic execution module is non-differentiable. For example, Neural Symbolic Machines~\cite{Liang2016NeuralSM} combines REINFORCE with a maximum-likelihood training process to find good programs. \citet{Guu2017FromLT} augment reinforcement learning with the maximum marginal likelihood so that probability is distributed evenly across consistent programs. Memory Augmented Policy Optimization (MAPO)~\cite{Liang2018MemoryAP} formulates its learning objective as an expectation over a memory buffer of high-reward samples and a separate expectation outside the buffer, which helps accelerate and stabilize policy gradient training. Meta Reward Learning~\cite{Agarwal2019LearningTG} uses an auxiliary reward function to provide feedback beyond a binary success or failure. Since these methods can only learn from sparse successful samples, they suffer from cold start and inefficient exploration of large search spaces. Recently, \citet{dai2017combining}, \citet{dai2019bridging}, and \citet{zhou2019abductive} introduce abductive learning, which states that human misperceptions can be corrected via abductive reasoning. In this paper, we follow the abductive learning method~\cite{li2020ngs} and propose a novel fixing mechanism to learn from negative samples, significantly accelerating and stabilizing the weakly-supervised learning process. We further design the tree regularization and memory buffer techniques to efficiently shrink and explore the solution space.
\section{Weakly-Supervised MWPs}
\begin{figure*}[htbp]
    \centering
    \includegraphics[width=\textwidth]{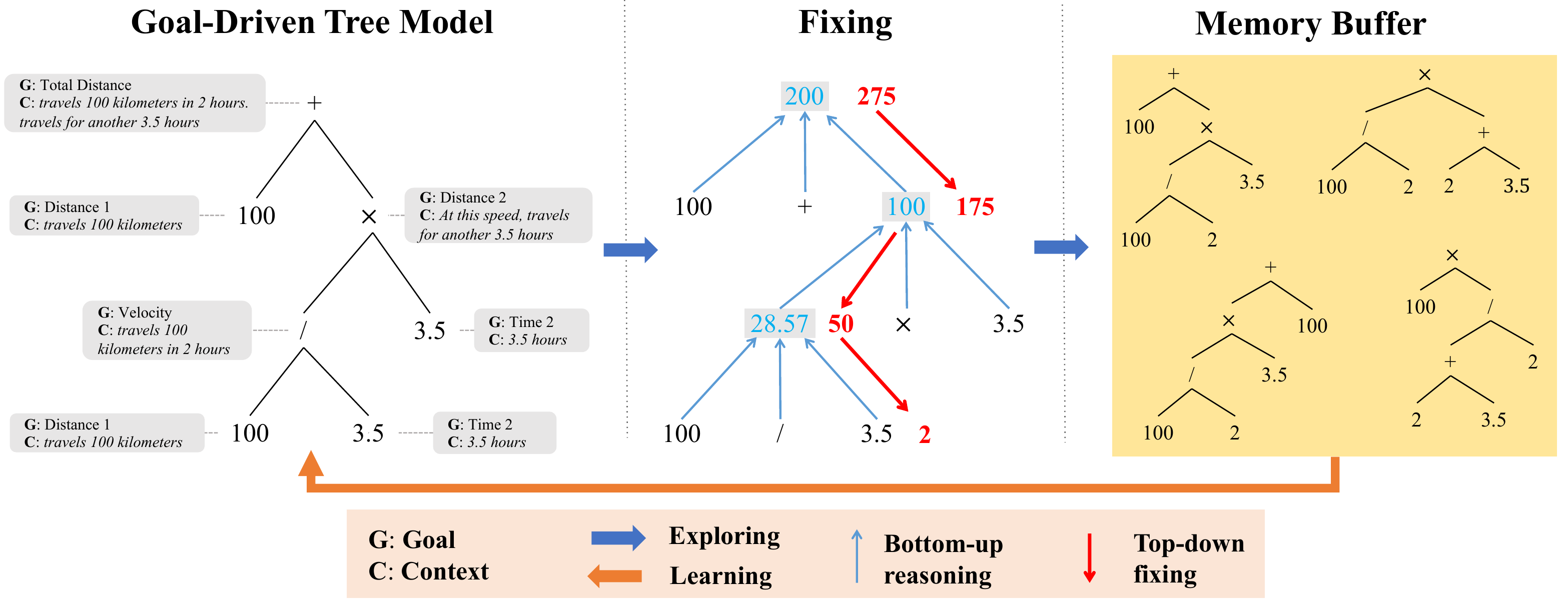}
    \caption{Overview of our proposed learning-by-fixing (LBF) method. It shows the process for learning the example in \autoref{fig:example}. LBF works by iteratively exploring the solution space and learning the MWP solver. \textbf{Exploring}: the problem first goes through the GTS module and produces a tentative solution using tree regularization. Then the fixing mechanism diagnoses this solution by propagating the correct answer in a top-down manner. The fixed solution is then added to the memory buffer. \textbf{Learning}: all solutions in the memory buffer are used as pseudo labels to train the GTS module using a cross-entropy loss function.}
    \label{fig:framework}
    \vskip -0.1in
\end{figure*}


In this section, we define the weakly-supervised math word problems and describe the goal-driven tree model originated from \citet{Xie2019AGT}.  Then we introduce the proposed learning-by-fixing method, as also shown in \autoref{fig:framework}.

\subsection{Problem Definition}
A math word problem is represented by an input problem text $P$. The machine learning model with parameters $\theta$ requires to translate $ P $ into an intermediate expression $ T $, which is executed to compute the final answer $y$.
In fully-supervised learning, we learn from the ground truth expression $T$ and the final answer $y$. The learning objective is to maximize the data likelihood $p(T,y|P; \theta) = p_\theta(T|P)p(y|T)$, where computing $y$ given $T$ is a deterministic process. In contrast, in the weakly-supervised setting, only $P$ and $y$ are observed, while $T$ is hidden. In other words, the model is required to generate an unknown expression from the problem text. The expression is then executed to get the final answer.


\subsection{Goal-driven Tree-Structured Model} \label{sec:GTS}
A problem text $P$ consists of words and numeric values. The model takes in problem text $P$ and generates a solution tree $T$. Let $V^{num}$ denote the ordered list of numeric values in
$P$ according to their order in the problem text. Generally, $T$ may contain constants $V^{con} = \{1,2,\pi\}$, mathematical operators $ V^{op} = \{+,-,\times,\div,\wedge\}$, and numeric values $V^{num}$ from the problem text $P$. Therefore, the target vocabulary of $P$ is denoted as $\Sigma = V^{op} \cup V^{con} \cup V^{num}$ and it varies between problems due to different $V^{num}$.
 
To generate the solution tree, we adopt the goal-driven tree-structured neural model (GTS)~\cite{Xie2019AGT}, which first encodes the problem text into its goal and then recursively decomposes it into sub-goals in a top-down manner. 



\noindent \textbf{Problem Encoding.} 
Each word of the problem text is encoded into a contextual representation. Specifically, for a problem $P = w_1w_2...w_n$, each word $w_i$ is first converted to a word embedding $\textbf{w}_i$. Then the sequence of embeddings is inputted to a bi-directional GRU~\cite{cho2014learning} to produce a contextual word representation: 
$\textbf{h}_i = \overrightarrow{\textbf{h}_i} + \overleftarrow{\textbf{h}_i},$
where $\overrightarrow{\textbf{h}_i}, \overleftarrow{\textbf{h}_i}$ are the hidden states of the forward and backward GRUs at position $i$, respectively. 

\noindent \textbf{Solution Tree Generation.}
The tree generation process is designed as a preorder tree traversal (root-left-right).
The root node of the solution tree is initialized with a goal vector $\textbf{q}_0 = \overrightarrow{\textbf{h}_n} +\overleftarrow{\textbf{h}_0}.$

For a node with goal $\textbf{q}$, we first derive a context vector $\textbf{c}$ by an attention mechanism to summarize relevant information from the problem:
\begin{align}
&a_i = \text{softmax}(\textbf{v}_a^\top \text{tanh}(\textbf{W}_a[\textbf{q},\textbf{h}_i])) \\
&\textbf{c} = \sum_i a_i \textbf{h}_i   
\end{align}
where $\textbf{v}_a$ and $\textbf{W}_a$ are trainable parameters. Then the goal $\textbf{q}$ and the context $\textbf{c}$ are used to predict the token of this node from the target vocabulary $\Sigma$. The probability of token $t$ is defined as:
\begin{align}
   &s(t|\textbf{q},\textbf{c}) = \textbf{w}_n^{\top}\text{tanh}(\textbf{W}_s[\textbf{q},\textbf{c},\textbf{e}(t)]) \label{eqn:eqn1}\\
   &p(t|\textbf{q},\textbf{c}) = \text{softmax}(s(t|\textbf{q},\textbf{c}))
   \label{eqn:eqn2}
\end{align}
where  $\textbf{e}(t)$ is the embedding of token $t$:

    \begin{equation}
    \textbf{e}(t)=
    \begin{cases}
      \textbf{M}_{op}(t) & \text{if}\ t\in V^{op} \\
      \textbf{M}_{con}(t) & \text{if}\ t\in V^{con}\\
       {\textbf{h}_{loc(t,P)}} & \text{if}\ t\in V^{num}
    \end{cases}
  \end{equation}
where $\textbf{M}_{op}$ and $\textbf{M}_{con}$ are two trainable embeddings for operators and constants, respectively. For a number token, its embedding is the corresponding hidden state ${\textbf{h}_{loc(t,P)}}$ from the encoder, where ${loc}(t,P)$ is the index of $t$ in the problem $P$. The predicted token $\hat{t}$ is:
\begin{equation}
\label{eq:token}
\hat{t} = \arg \max_{t \in \Sigma} p(t|\textbf{q},\textbf{c})
\end{equation}
If the predicted token is a number token or constant, the node is terminated and its goal is realized by the predicted token; otherwise, the predicted token is an operator and the current goal is decomposed into left and right sub-goals combined by the operator. Please refer to the \textit{supplementary material} for more details about the goal decomposition process.

\noindent \textbf{Answer Calculation.} The generated solution tree is transformed into a reasoning tree $\hat{T}$ by creating auxiliary non-terminal nodes in place of the operator nodes to store the intermediate results, and the original operator nodes are attached as child nodes to the corresponding auxiliary nodes. Then the final answer $\hat{y}$ is calculated by executing $\hat{T}$ to the value of the root node in a bottom-up manner.

\subsection{Learning-by-Fixing}
\subsubsection{Fixing Mechanism} \label{sec:fix}
Drawing inspiration from humans' ability to correct and learn from failures, we propose a fixing mechanism to correct the wrong solution trees via abductive reasoning following \citet{li2020ngs} and use the fixed solution trees as pseudo labels for training. Specifically, we find the most probable fix for the wrong prediction by back-tracking the reasoning tree and propagating the error from the root node into the leaf nodes in a top-down manner.

The key ingredient in the fixing mechanism is the 1-step fix (1-FIX) algorithm which assumes that only one symbol in the reasoning tree can be substituted. As shown by the $\textsc{1-Fix}$ function in \autoref{alg_m_fix}, the 1-step fix starts from the root node of the reasoning tree and gradually searches down to find a fix that makes the final output equal to the ground-truth. The search process is implemented with a priority queue, where each element is defined as a fix-tuple $(A, \alpha_A, p)$: 
\begin{itemize}[leftmargin=*,noitemsep]
    \item $A$ is the current visiting node.
    \item $\alpha_A$ is the expected value on this node, which means if the value of $A$ is changed to $\alpha_A$, $\hat{T}$ will execute to the ground-truth answer $y$.
    \item $p$ is the visiting priority, which reflects the probability of changing the value of $A$. 
\end{itemize}


In 1-FIX, error propagation through the solution tree is achieved by a $solve$ function, which aims at computing the expected value of a child node from its parent's expected value. Supposing $B$ is $A$'s child node and $\alpha_A$ is the expected value of $A$, the $solve(B,A,\alpha_A)$ function works as following:

\begin{itemize}[leftmargin=*,noitemsep]
    \item If $B$ is $A$'s left or right child, we directly solve the equation $\alpha_B \bigoplus child_{R}(A) = \alpha_A$ or $child_{L}(A) \bigoplus \alpha_B = \alpha_A$ to get $B$'s expected value $\alpha_B$, where $\bigoplus$ denotes the operator.
    \item If $B$ is an operator node, we try to replace $B$ with all other operators and check whether the new expression can generate the correct answer. That is, $child_{L}(A)\  
    \alpha_B\  child_R(A) = \alpha_A$ where $\alpha_B$ is now an operator. If there is no $\alpha_B$ satisfying this equation, the solve function returns none.
\end{itemize}
Please refer to the \textit{supplementary material} for the definition of the visiting priority as well as the illustrative example of the 1-FIX process.

To search the neighbors of $\hat{T}$ within multi-step distance, we extend the 1-step fix to multi-step by incorporating a $\textsc{RandomWalk}$ function. As shown in \autoref{alg_m_fix}, if we find a fix by 1-FIX, we return this fix; otherwise, we randomly change one leaf node in the reasoning tree to another symbol within the same set (\eg, operators $V^{op}$) based on the probability in \autoref{eqn:eqn2}. This process will be repeated for certain iterations until it finds a fix for the solution.
\begin{algorithm}[h]
\small
    \caption{Fixing Mechanism} \label{alg_m_fix}
    \label{alg:training}
    \begin{algorithmic}[1]
    \STATE \textbf{Input}: $\text{reasoning tree}~\hat{T}, \text{ground-truth answer}~y$
    \STATE $T^{(0)} = \hat{T}$
    \FOR{$i \gets 0~\text{to}~m$}
        \STATE $T^* = \textsc{1-Fix}(T^{(i)}, y)$
        \IF {$T^* \neq \varnothing$}
            \STATE \textbf{return} $T^*$
        \ELSE
            \STATE $T^{(i+1)} = \textsc{RandomWalk}(T^{(i)})$
        \ENDIF
    \ENDFOR
    \STATE \textbf{return} $\varnothing$
    \STATE
    
    \STATE \textbf{function} $\textsc{1-Fix}(T, y)$
    \STATE $q$ = PriorityQueue(), $S$ = the root node of $T$
    \STATE $q.push(S, y, 1)$ 
    \WHILE{$(A, \alpha_A, p) = q.pop()$}
        \IF{$A \in \Sigma$}
            \STATE $T^* = \hat{T}(A \to \alpha_A)$
            \STATE \textbf{return} $T^*$
        \ENDIF
        \FOR{$B \in child(A)$}
            \STATE $\alpha_B = solve(B, A, \alpha_A)$
            \IF {not ($B \in \Sigma$ and $\alpha_B \notin \Sigma$)}
                \STATE $q.push(B, \alpha_B, p(B\to\alpha_B))$
            \ENDIF
        \ENDFOR
    \ENDWHILE
    \STATE \textbf{return} $\varnothing$
    
    \end{algorithmic}
\end{algorithm}

\subsubsection{Solution Space Exploration}
\hfill \break \textbf{Tree Regularization}
While \citet{li2020ngs} assumes the length of the intermediate representation is given, the expression length is unknown in weakly-supervised learning. Thus, the original solution space is infinite since the predicted token decides whether to continue the generation or stop. Therefore, it is critical to shrink the solution space, \ie, control the size of the generated solution trees. If the size of the generated solution tree varies a lot from the target size, it would be challenging for the solution or its fix to hit the correct answer. Although the target size is unknown, we observe a positive correlation between the target size and the number of quantities in text. Regarding this observation as a tree size prior, we design a tree regularization algorithm to generate a solution tree with a target size and regularize the size in an empirical range. Denote the size of a solution tree $\text{Size}(T)$ as the number of leaf nodes including quantities, constants, and operators. The prior range of $\text{Size}(T)$ given the length of the numeric value list $\text{len}(V^{num})$ is defined as:
\begin{equation} \label{eq_tree_reg}
\begin{split}
&\text{Size}(T) \in [\text{minSize}(T),\text{maxSize}(T)]\\
&\text{minSize}(T) = a_{min} \text{len}(V^{num}) + b_{min}\\
&\text{maxSize}(T) = a_{max} \text{len}(V^{num})+b_{max}
\end{split}
\end{equation}
where $a_{min}, b_{min}, a_{max}, b_{max}$ are the hyperparameters. The effect of these hyperparameters will be discussed in \autoref{tab:beam}. 

\begin{figure}[ht] 
    \centering 
    \includegraphics[width=\linewidth]{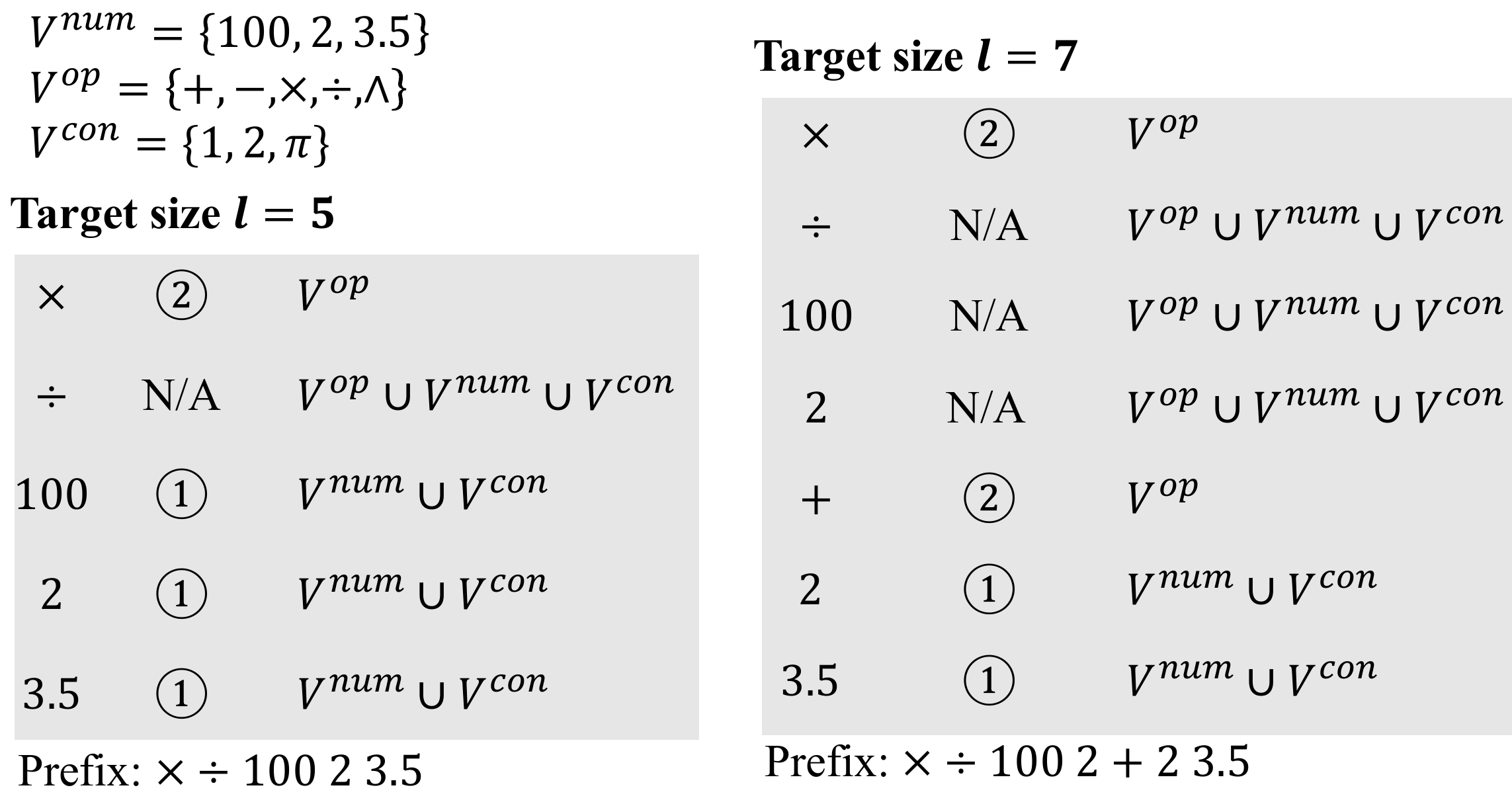} \vskip -0.05in
    \caption{Tree regularization for the problem in \autoref{fig:example} given different target sizes. The three columns are the generated tokens, the effective rules, and the target vocabularies shrunk by the rules, respectively. } \label{fig:tree_reg}
    \vspace{-5mm}
\end{figure}

We further propose a \textit{tree regularization} algorithm to decode a solution tree with a given size. To generate a tree of a given size $l$, we design two rules to produce a prefix-order expression during the preorder tree decoding:
\begin{enumerate}[leftmargin=*,noitemsep]
    \item The number of operators cannot be greater than $\floor{l/2}$.
    \item Except the $l$-th position, the number of numeric values (quantities and constants) cannot be greater than the number of operators.
\end{enumerate}
These two rules are inspired by the syntax of prefix notation (a.k.a, normal Polish notation) for mathematical expressions. The rules shrink the target vocabulary $\Sigma$ in \autoref{eq:token} so that the tree generation can be stopped when it reaches the target size. \autoref{fig:tree_reg} shows illustrative examples of the tree regularization algorithm.

With tree regularization, we can search the possible fixes within a given range of tree size $\left[\text{minSize}(T), \text{maxSize}(T)\right]$ for each problem. 

\noindent\textbf{Memory Buffer.} 
We adopt a memory buffer to track and save the discovered fixes for each problem. The memory buffer enables us to seek multiple solutions for a single problem and use all of them as pseudo labels for training, which encourages diverse solutions. Formally, given a problem $P$ and its buffer $\beta$, the learning objective is to minimize the negative log-likelihood of all fixed expressions in the buffer:
\begin{equation} \label{eq:loss}
J(P, \beta) =-\sum_{T^{\ast} \in \beta}\log p(T^{\ast}|P)
\end{equation}
\subsection{Learning-by-Fixing Framework}
The complete learning-by-fixing method is described in \autoref{alg:ma-fix}. In the exploring state, we use the fixing mechanism and tree regularization to discover possible fixes for the wrong trees generated by the neural network, and put them into a buffer. In the learning stage, we train the model with all the solutions in the memory buffer by minimizing the loss function in \autoref{eq:loss}.
\begin{algorithm}
\small
\begin{algorithmic}[1]
\caption{Learning-by-Fixing}\label{alg:ma-fix}
\STATE \textbf{Input}: training set $\mathcal{D}=\{(P_i, y_i)\}_{i=1}^N$
\STATE memory buffer $\mathcal{B}=\{\beta_i\}_{i=1}^N$, the GTS model $\theta$

\FOR {$P_i, y_i, \beta_i \in (\mathcal{D,B})$} 
    \STATE \hfill$\triangleright$\textit{Exploring}
    \STATE $\hat{T}_i$ = GTS ($P; \theta$)
    \STATE $T_i^{\ast} = m\text{-FIX}(\hat{T_i}, y_i$)
     \IF {$T_i^{\ast} \neq  \varnothing $ and $T_i^{\ast} \notin \beta_i$}
            \STATE $\beta_i \leftarrow \beta_i \cup \{T_i^{\ast}\}$
     \ENDIF
     
     \STATE \hfill$\triangleright$\textit{Learning}
     \STATE $\theta = \theta - \nabla_{\theta} J(P_i, \beta_i)$
\ENDFOR
 \end{algorithmic}

\end{algorithm}\\


\vspace{-7mm}
\section{Experimental Results} \label{sec:exp}
\subsection{Experimental Setup}
\noindent \textbf{Dataset.} We evaluate our proposed method on the Math23K dataset~\cite{wang-etal-2017-deep}. It contains 23,161 math word problems annotated with solution expressions and answers. For the weakly-supervised setting, we only use the problems and final answers and discard the expressions. We do cross-validation following the setting of \citet{Xie2019AGT}.


\noindent \textbf{Evaluation Metric.} We evaluate the model performance by answer accuracy, where the generated solution is considered correct if it executes to the ground-truth answer. Specifically, we report answer accuracies of all the top-$1/3/5$ predictions using beam search. It evaluates the model's ability to generate multiple possible solutions. 

\noindent \textbf{Models.} We conduct experiments by comparing our methods with variants of weakly-supervised learning methods. Specifically, we experiment with two inference models: Seq2Seq with bidirectional Long Short Memory network (BiLSTM)~\cite{Wu2016GooglesNM} and GTS~\cite{Xie2019AGT}, and train with four learning strategies: REINFORCE, MAPO~\cite{Liang2018MemoryAP}, LBF, LBF-w/o-M (without memory buffer). MAPO is a state-of-the-art method in semantic parsing task that extends the REINFORCE with augmented memory. Both models are also trained with the tree regularization algorithm. We also compare with the fully-supervised learning methods to demonstrate our superiority in generating diverse solutions. In the ablative studies, we analyze the effect of the proposed tree regularization and the length of search steps in fixing mechanism. 

\subsection{Comparisons with State-of-the-art}
\autoref{tab:acc} summarizes the answer accuracy of different weakly-supervised learning methods and the state-of-the-art fully-supervised approaches. The proposed learning-by-fixing framework significantly outperforms the policy gradient baselines like REINFORCE and MAPO, on both the Seq2seq and the GTS models. It demonstrates the strength of our proposed LBF method in weakly-supervised learning.  The GTS-LBF-fully model is trained by initializing the memory buffer with all the ground-truth expressions. It demonstrates that by extending to the fully-supervised setting, our model maintains the top-1 accuracy while significantly improving solutions' diversity. We believe that learning MWPs with weak supervision is a promising direction. It requires fewer annotations and allows us to build larger datasets with less cost.

\begin{table}[h]
    \centering
    \small
    \scalebox{0.75}{
    \begin{tabular}{c|c|c}
        \hline
        \multicolumn{2}{c|}
        {\textbf{Model}} & \textbf{Accuracy}(\%) \\
        \hline
        \multicolumn{3}{c}{\textit{Fully-Supervised}} \\
        \hline
        \multicolumn{2}{c|}{{Retrieval}~\cite{Robaidek2018DataDrivenMF}}  & 47.2\\
        \multicolumn{2}{c|} {Classification~\cite{Robaidek2018DataDrivenMF}} & 57.9\\
        
        \multicolumn{2}{c|}{ LSTM~\cite{Robaidek2018DataDrivenMF}} & 51.9\\
        
        \multicolumn{2}{c|}{
        CNN~\cite{Robaidek2018DataDrivenMF}} & 42.3\\
        
        \multicolumn{2}{c|}{
        DNS~\cite{wang-etal-2017-deep}} & 58.1\\
        
         \multicolumn{2}{c|} {Seq2seqET~\cite{Wang2018TranslatingMW}} & 66.7\\
        
         \multicolumn{2}{c|}
         {Stack-Decoder~\cite{Chiang2019SemanticallyAlignedEG}} & 65.8\\
         
          \multicolumn{2}{c|}
         {T-RNN~\cite{Wang_Zhang_Zhang_Xu_Gao_Dai_Shen_2019}} & 66.9\\
         
          \multicolumn{2}{c|}
         {GTS~\cite{Xie2019AGT}} & 74.3 \\
         \multicolumn{2}{c|}
         {Graph2Tree~\cite{zhang2020graph2tree}} & \textbf{74.8} \footnotemark\\
         \multicolumn{2}{c|}
         {GTS-LBF-fully}  & 74.1 \\
        \hline

        \multicolumn{3}{c}{\textit{Weakly-Supervised}} \\
        \hline
        \multirow{4}{*}{Seq2seq} & REINFORCE & 1.2\\
        & MAPO & 10.7\\
        & LBF-w/o-M & 44.7\\
        & LBF & 43.6\\
        \hline
        \multirow{4}{*}{GTS} & REINFORCE & 15.8\\
         & MAPO & 20.8 \\
         & LBF-w/o-M & 58.3 \\
         & LBF  & \textbf{59.4}\\
         \hline
    \end{tabular}}
    \caption{Answer accuracy on the Math23K dataset. We compare variants of models with our LBF method.}
    \label{tab:acc}
    \vspace{-5mm}
\end{table}

\footnotetext{
We run the code using the same setting as GTS for three times and compute the average accuracy.}
\subsection{Convergence Speed}
\begin{figure}[htbp]
    \centering
    \includegraphics[width=\linewidth]{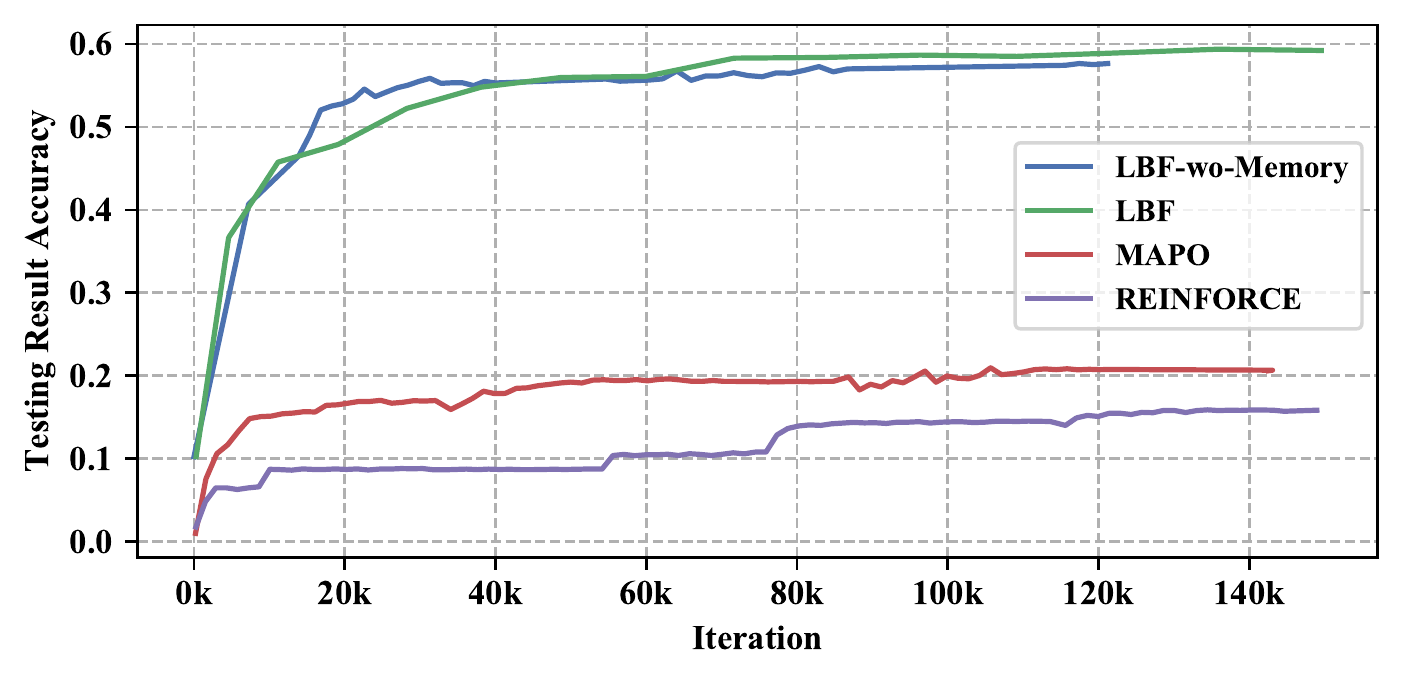}
    \caption{The learning curves of the GTS model using different weakly-supervised learning methods.}
    \label{fig:curve}
    \vspace{-5mm}
\end{figure}

\autoref{fig:curve} shows the learning curves of different weakly-supervised learning methods for the GTS model. The proposed LBF method converges significantly faster and achieves higher accuracy compared with other methods. Both the REINFORCE and MAPO take a long time to start improving, which indicates the policy gradient methods suffer from the cold-start and need time to accumulate rewarding samples. 

\subsection{Diverse Solutions with Memory Buffer}
To evaluate the ability to generate diverse solutions, we report the answer accuracies of all the top-1/3/5 solutions on the test set using beam search, denoted as Acc@1/3/5, as shown in \autoref{tab:beam}. In the weakly-supervised scenario, GTS-LBF achieves slightly better Acc@1 accuracy and much better Acc@3/5 accuracy than GTS-LBF-w/o-M. In the fully supervised scenario, GTS-LBF-fully achieves comparable Acc@1 accuracy and much better Acc@3/5 accuracy than the original GTS model. Particularly, GTS-LBF-fully outperforms GTS by 21\% and 26\% in terms of Acc@3/5 accuracy. It reveals the efficacy of the memory buffer in encouraging diverse solutions in both weakly-supervised learning and fully-supervised learning.

\begin{table}[!htbp]
    \centering
    \small
    \begin{tabular}{c|l|rrr}
        \hline
        \textbf{Model} & \textbf{Tree Size} &\textbf{Acc@1} &\textbf{Acc@3} & \textbf{Acc@5} \\
        \hline
        \multicolumn{5}{c}{\textit{Fully Supervised}}\\
        \hline
        \multicolumn{2}{c|}{GTS}  & \textbf{74.3} & 42.2 & 30.0\\
        \hline
        \multicolumn{2}{c|} {GTS-LBF-fully} & 74.1 & \textbf{63.4} & \textbf{56.3} \\
        \hline
        \multicolumn{5}{c}{\textit{Weakly Supervised}}\\
        \hline
        \multirow{4}{*}{\shortstack{GTS-LBF-\\w/o-M}} 
        & [1,$+\infty$) & $\sim$0 & $\sim$0 & $\sim$0\\
        & [2n-1,2n+1] & 55.3 & 26.2 & 19.3\\
        & [2n-1,2n+3] & 58.3 & 27.7 & 20.3\\
        & [2n-3,2n+5] & 56.7 & 27.7 & 20.6\\
        \hline
        \multirow{4}{*}{GTS-LBF} 
        & [1,$+\infty$) & $\sim$0 & $\sim$0 & $\sim$0\\
        & [2n-1,2n+1] & 56.7 & 45.3 & 39.1\\
        & [2n-1,2n+3] & \textbf{59.4} & \textbf{49.6} & \textbf{45.2}\\
        & [2n-3,2n+5] & 57.6 & 49.3 & 45.2\\
        \hline
    \end{tabular}
    \caption{Answer accuracies of all the top-1/3/5 solutions decoded using beam search, denoted as Acc@1/3/5.}
    \label{tab:beam}
\vspace{-5mm}

\end{table}

\begin{figure*}[htbp]
    \vspace{-0.5cm}
    \centering
    \includegraphics[width=\textwidth, page=1]{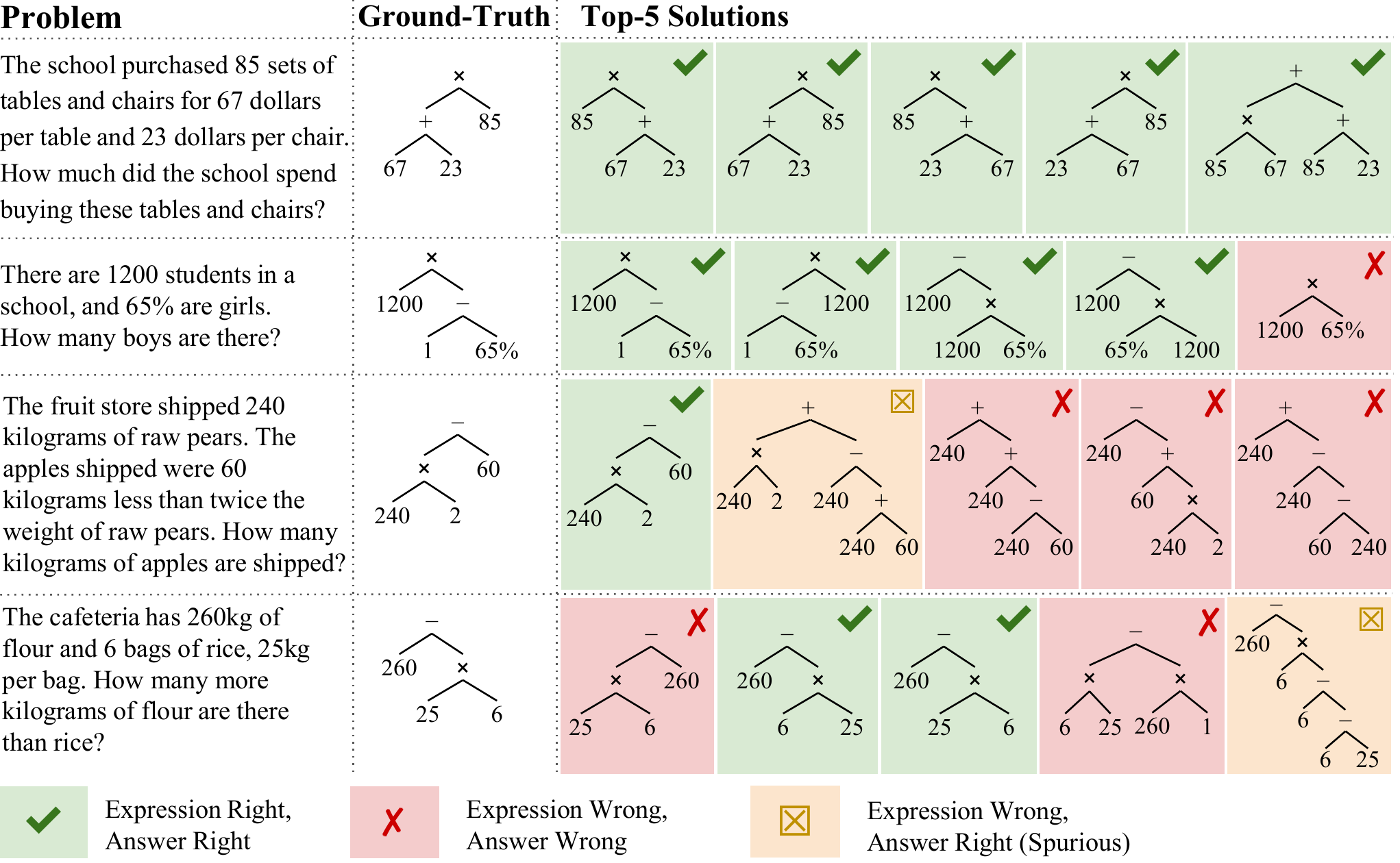}
    \vspace{-0.5cm}
    \caption{Qualitative results on the Math23K dataset. We visualize the solution trees generated by our method.}
    \vspace{-0.5cm}

    \label{fig:qualitative}
\end{figure*}

\subsection{Qualitative Analysis} \label{sec_QS}
We visualize several examples of the top-5 predictions of GTS-LBF in \autoref{fig:qualitative}. In the first example, the first solution generated by our model is to sum up the prices of a table and a chair first, and then multiply it by the number of pairs of tables and chairs. Our model can also produce another reasonable solution (the fifth column) by deriving the prices of tables and chairs separately and then summing them up. 


One caveat for the multiple solutions is that some solutions have different solution trees but are equivalent by switching the order of numeric values or subtrees, as shown in the first four solutions of the first problem in \autoref{fig:qualitative}. In particular, multiplication and addition are commutative, and our model learns and exploits this property to generate equivalent solutions with different tree structures. 
\begin{table}[]
    \centering
    \small
    \begin{tabular}{c|ccc}
        \hline
         & Right & Wrong & Spurious \\
         \hline
        Acc@1 & 58.6 & 40.6 & 0.56\\
        Acc@3 & 49.3 & 50.4 & 0.27\\
        Acc@5 & 44.9 & 54.8 & 0.32\\
        \hline
    \end{tabular}
    \caption{Human evaluation on the generated solutions (\%). }
    \label{tab:spurious}
    \vspace{-5mm}
\end{table}

The first solution to the fourth problem in \autoref{fig:qualitative} is a typical error case of our model due to the wrong prediction of the problem goal. Another failure type is the spurious solutions, which are correct but not meaningful answers, such as the second solution of the third problem in \autoref{fig:qualitative}. To test how frequent the spurious solutions appear, we randomly select 500 examples from the test set, and ask three human annotators to determine whether each generated expression is right, wrong, or spurious. \autoref{tab:spurious} provides the human evaluation results, and it shows that spurious solutions are rare in our model. 

\subsection{Ablative Analyses}
\paragraph{Tree Regularization.}
We test different choices of the hyperparameters defined by \autoref{eq_tree_reg} in tree regularization. As shown in \autoref{tab:beam}, the model without tree regularization, \ie, tree size $\in [1, +\infty)$, fails to converge and gets nearly 0 accuracy. The best range for the solution tree size is $[2n-1, 2n+3]$, where $n = \text{len}(V^{num})$.  We provide an intuitive interpretation of this range: for a problem with $n$ quantities, (1) $n-1$ operators are needed to connect $n$ quantities, which leads to the lower bound of tree size to $2n-1$; (2) in certain cases, the constants or quantities are used more than once, leading to a rough upper bound of $2n+3$. Therefore, we use $[2n-1, 2n+3]$ as the default range in our implementations.
Empirically, this range covers 88\% of the lengths of the given ground-truth expressions in the Math23K dataset, providing an efficient prior for tree size.

\vspace{-3mm}

\paragraph{Number of Search Steps} 
\autoref{tab:step} shows the comparison of various step lengths in the m-FIX algorithm. In most cases, increasing the step length improves the chances of correcting wrong solutions, thus improving the performance.
\begin{table}[H]
    \centering
    \vspace{-3mm}
    \small{
    \begin{tabular}{c|cccc}
        \hline
        \backslashbox[33mm]{\textbf{Models}}{\textbf{Steps}} & \textbf{1} & \textbf{10} & \textbf{50 (default)} & \textbf{100}\\
        \hline
        Seq2seq-LBF-w/o-M & 41.9 & 43.4 & {44.7} & \textbf{47.8}\\
        Seq2seq-LBF &43.9 &\textbf{45.7}&43.6 & 44.6\\
        \hline
        GTS-LBF-w/o-M & 51.2 & 54.6 & \textbf{58.3} & 57.8\\
        GTS-LBF &52.5 & 55.8  & 59.4 & \textbf{59.6}\\
        \hline
    \end{tabular}
    }
    \caption{Accuracy (\%) using various search steps.}
    \label{tab:step}
 \vspace{-3mm}
   
\end{table}

\section{Conclusion}
In this work, we propose a weakly-supervised paradigm for learning MWPs and a novel learning-by-fixing framework to boost the learning.
Our method endows the MWP learner with the capability of learning from wrong solutions, thus significantly improving the answer accuracy and learning efficiency.
One future direction of the proposed model is to prevent generating equivalent or spurious solutions during training, possibly by making the generated solution trees more interpretable with semantic constraints.

\section{Ethical Impact}
The presented work should be categorized as research in the field of weakly-supervised learning and abductive reasoning. It can help teachers in school get various solutions of a math word problem. This work may also inspire new algorithmic, theoretical, and experimental investigation in neural-symbolic methods and NLP tasks.

\section{Acknowledgement}
This work reported herein is supported by ARO W911NF1810296, DARPA XAI N66001-17-2-4029, and ONR MURI N00014-16-1-2007. 

\bibliography{emnlp2020}
\end{document}